\documentclass[runningheads]{llncs}

\usepackage{eccv}

\usepackage{eccvabbrv}

\usepackage{graphicx}
\usepackage{booktabs}

\usepackage[accsupp]{axessibility}  

\usepackage{hyperref}

\usepackage{orcidlink}
\usepackage{enumerate}
\usepackage{enumitem}
\usepackage{adjustbox}
\usepackage{graphicx}
\usepackage{caption}
\usepackage{makecell}
\usepackage{array}
\usepackage{booktabs}
\usepackage{amssymb}
\usepackage{amsmath}
\usepackage{amsfonts}  
\usepackage{bm}  
\usepackage{pifont}
\usepackage{wasysym}
\usepackage{siunitx}
\usepackage{adjustbox}
\usepackage{multirow}
\usepackage{color}
\usepackage{caption}
\usepackage{float} 
\usepackage{subfiles}
\usepackage{setspace}
\usepackage{makecell}
\usepackage{colortbl}
\usepackage{diagbox}
\usepackage{marvosym}

\begin{document}

\title{Long-range Turbulence Mitigation: A  Large-\\scale Dataset and A Coarse-to-fine Framework} 

\titlerunning{Long-range Turbulence Mitigation}

\author{Shengqi Xu \and
Run Sun \and
Yi Chang \textsuperscript{\Letter}  \and
Shuning Cao \and
Xueyao Xiao \and
Luxin Yan }

\authorrunning{Xu et al.}

\institute{National Key Lab of Multispectral Information Intelligent Processing Technology, \\
Huazhong University of Science and Technology, China
\email{\{shengqi,sunrun,yichang,caoshuning,xiaoxueyao,yanluxin\}@hust.edu.cn}\\
\url{https://shengqi77.github.io/RLR-AT.github.io/}}


\onecolumn{%
\renewcommand\onecolumn[1][]{#1}%
\maketitle

\begin{center}
  \captionsetup{skip=0.8mm} 
    \centering
    \captionsetup{type=figure}
    \includegraphics[width=0.84\textwidth]{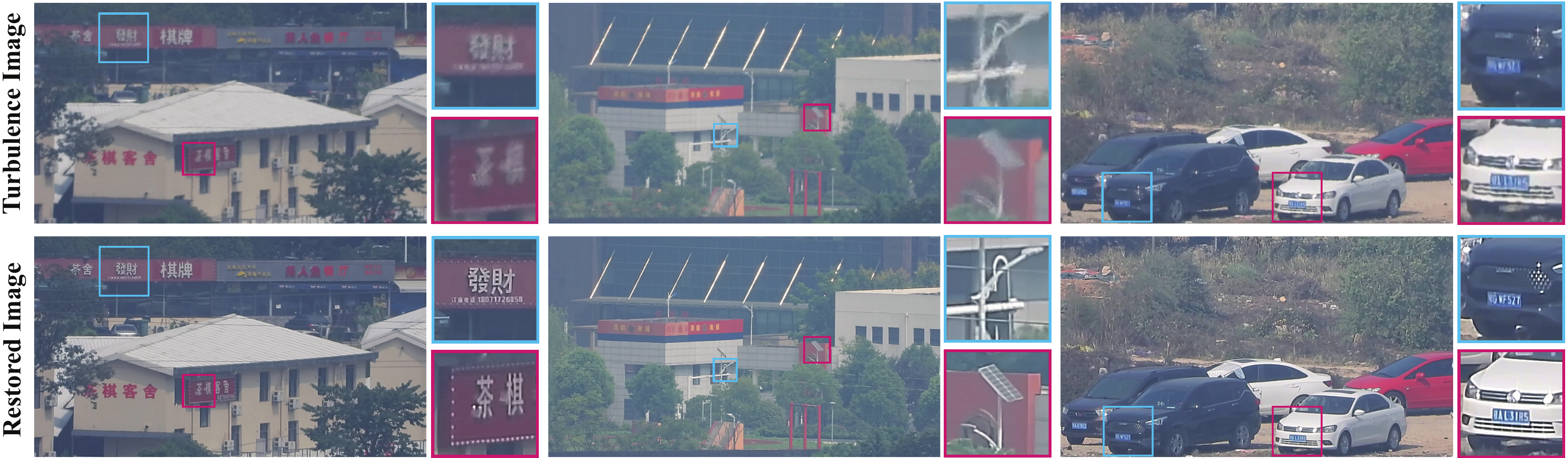}
    \captionof{figure}{Visual examples of turbulence mitigation on proposed real-world long-range atmospheric turbulence benchmark RLR-AT. The proposed method could effectively handle the long-range turbulence with \textbf{severe distortions}.}
 \label{Figure1}
\end{center}%
}

\begin{abstract} 
Long-range imaging inevitably suffers from atmospheric turbulence with severe geometric distortions due to random refraction of light. The further the distance, the more severe the disturbance. Despite existing research has achieved great progress in tackling short-range turbulence, there is less attention paid to long-range turbulence with significant distortions. To address this dilemma and advance the field, we construct a large-scale real long-range atmospheric turbulence dataset (RLR-AT), including 1500 turbulence sequences spanning distances from 1 Km to 13 Km. The advantages of RLR-AT compared to existing ones: turbulence with longer-distances and higher-diversity, scenes with greater-variety and larger-scale. Moreover, most existing work adopts either registration-based or decomposition-based methods to address distortions through one-step mitigation. However, they fail to effectively handle long-range turbulence due to its significant pixel displacements. In this work, we propose a coarse-to-fine framework to handle severe distortions, which cooperates dynamic turbulence and static background priors (CDSP).  On the one hand, we discover the pixel motion statistical prior of turbulence, and propose a frequency-aware reference frame for better large-scale distortion registration, greatly reducing the burden of refinement. On the other hand, we take advantage of the static prior of background, and propose a subspace-based low-rank tensor refinement model to eliminate the misalignments inevitably left by registration  while well preserving details. The dynamic and static priors complement to each other, facilitating us to progressively mitigate long-range turbulence with severe distortions. Extensive experiments demonstrate that the proposed method outperforms SOTA methods on different datasets.
\end{abstract}

\section{Introduction}
\label{sec:intro}
Seeing farther and more clearly is crucial for many military and civilian applications. Unfortunately, long-range imaging inevitably suffers from the atmospheric turbulence with severe geometric distortion due to random refraction of light \cite{tatarski2016wave,chan2022tilt,fried1966optical}. When accumulating over long distance in a non-uniform atmosphere medium, more refractions occur during optical transmission, resulting in more severe pixel displacement. Although short-range atmospheric turbulence mitigation has achieved great progress in recent years \cite{jin2021neutralizing,mao2022single,yasarla2021learning,zhu2012removing,he2016atmospheric,hua2020removing}, very few studies have focused on long-range turbulence. \textit{In this work, our goal is to handle the long-range turbulence with severe distortions as shown in Fig. \ref{Figure1}.}

Atmospheric turbulence benchmark is a key issue for evaluating turbulence mitigation methods. Existing datasets can be classified into synthetic datasets \cite{zhang2022imaging,zhang2024spatio} and real-world datasets \cite{hirsch2010efficient,Textdataset,jin2021neutralizing,anantrasirichai2013atmospheric,gilles2017open,oreifej2012simultaneous,mao2022single}. Synthetic datasets are mainly constructed using turbulence simulators \cite{mao2020image,chimitt2020simulating,chimitt2022real}. However, simulators cannot precisely generate data entirely matching the features of real turbulence, since turbulence in real scenarios possesses complex nature, especially for long-range turbulence. Real turbulence is primarily influenced by temperature and imaging distance. Greater distances and higher temperatures both lead to more severe turbulence. Typically, real turbulence can be divided into: hot-air turbulence and long-range turbulence \cite{mao2020image}. Most public real datasets consist of hot-air turbulence, and TurbRecon \cite{mao2020image} stands out by capturing 2 turbulence sequences for building scene at a distance of 4 Km. As such, constructing a large-scale real-world long-range turbulence dataset with diverse scenes is highly necessary.

In this work, we construct a large-scale real-world long-range turbulence dataset RLR-AT for long-range turbulence mitigation. The strength of our benchmark is threefold. Firstly, RLR-AT contains long-range turbulence with longer-distances and higher-diversity, covering diverse distortions ranging from 1 Km to 13 Km. Secondly, it consists of large-scale and diverse scenes, including 1500 turbulence sequences collected across various scenarios, such as text, object, building, etc. Last but not least, RLR-AT is collected by a telephoto camera with high-resolution (1980*1080 pixels). Overall, RLR-AT can serve as a benchmark for future works targeting long-range turbulence mitigation.

The main difficulty of long-range turbulence mitigation lies in the severe distortions. To handle distortions, most existing methods can be classified into two categories: registration-based methods \cite{caliskan2014atmospheric,fazlali2022atmospheric,hua2020removing,mao2020image,shimizu2008super,zhu2012removing,xie2016removing} and decomposition-based methods \cite{deshmukh2013fast,oreifej2012simultaneous,he2016atmospheric}. The former merely employs a zero-mean assumption prior of dynamic turbulence to construct a reference frame, aligning the distorted frames with the reference frame using registration technique. However, such strategy struggles to handle long-range turbulence with severe distortions, as registration errors emerge from the blurring of the reference frame.

On the contrary, the latter is to treat the distortions among the frames as gross error, and directly remove the distortions through matrix decomposition by exploring low-rank prior of static background. Though it is effective for mild distortions, it is theoretically less robust when handling severely corrupted observations \cite{wright2009robust,lu2016tensor}. Thus, directly employing such strategy on the long-range turbulence with severe distortions would suffer from unexpected details loss. Overall, previous methods either utilize the zero-mean assumption of dynamic turbulence or low-rank prior of static background, and both of them are difficult to handle long-range turbulence with severe distortions.

To handle long-range turbulence with severe distortions, we propose a coarse-to-fine framework that cooperates dynamic turbulence and static background priors (CDSP). On the one hand, we explore the pixel motion statistical prior of turbulence and discover that the \textit{pixel occurring most frequently at one certain position is most likely closer to the original GT.} This inspires us to propose a frequency-aware reference frame for better distortion registration, significantly reducing the burden of subsequent refinement.  On the other hand, we further take advantage of static prior of background  and  propose a subspace-based low-rank tensor refinement model to refine the registration errors unavoidably left by registration meanwhile well preserving details.
The dynamic and static priors complement to each other, facilitating us to mitigate long-range turbulence with severe distortions. Overall, our main contributions are summarized as follows:

\begin{enumerate}[leftmargin=10pt]
  \item  Our work focuses on a challenging yet practical task: long-range turbulence mitigation. We construct a large-scale real-world long-range atmospheric turbulence benchmark (RLR-AT). Compared to existing public real-world datasets , RLR-AT is the farthest (ranging from 1 Km to 13 Km) and  largest-scale (1500 sequences with high-resolution 1980*1080) turbulence dataset with diverse turbulence levels and scenes.  This dataset would be a good testbed for the community, especially for long-range turbulence mitigation.
  \setlength{\itemsep}{0pt}
  \item We propose a coarse-to-fine framework for long-range turbulence mitigation, which cooperates dynamic and static priors. Specifically,  we figure out the pixel displacement statistical prior of dynamic turbulence and propose a frequency-aware reference frame for better registration, significantly reducing the burden of refinement. Moreover, we take advantage of low-rank prior of static background and propose a subspace-based low-rank tensor refinement model to remove the registration errors meanwhile well preserving details. Compared to existing methods, the dynamic and static priors complement to each other, facilitating us to address long-range turbulence with significant distortions.
  \setlength{\itemsep}{0pt}
  \item We comprehensively compare CDSP with existing methods on proposed real long-range turbulence dataset RLR-AT and synthetic dataset. Extensive experiments show that our CDSP consistently outperforms SOTA methods, especially when handling long-range turbulence with severe distortions.
\end{enumerate}

\section{Related Work}

\textbf{Real Atmospheric Turbulence Benchmarks}. Atmospheric turbulence is a fundamental issue in long-range imaging system, mostly depending on the temperature and imaging distance. Higher temperatures and longer distances typically result in more severe turbulence. Generally, turbulence can be mainly classified into two categories: hot-air turbulence and long-range turbulence \cite{mao2020image}. In Table \ref{tab_datasets}, we provide a comprehensive summary of existing pubic benchmarks. EEF \cite{hirsch2010efficient} provided two widely used hot-air turbulence samples. UG2+ TurbuText \cite{Textdataset} and Heat Chamber \cite{mao2022single} collected turbulence sequences with a distance of 20 meters. These hot-air turbulence datasets were collected with artificial heat burner. Further, CLEAR \cite{anantrasirichai2013atmospheric}, OTIS \cite{gilles2017open}, Turbulence Text \cite{mao2022single} and TSR-WGAN \cite{jin2021neutralizing} collected turbulence under high-temperature environment near the surface. Most existing  public datasets are composed of hot-air turbulence, and TurbRecon \cite{mao2020image} captured 2 turbulence sequences with distance of 4Km. Moreover, existing datasets are still limited in terms of scale and diversity. In this work, we focus on the challenging problem: long-range turbulence mitigation and construct a large-scale real long-range turbulence dataset to advance this field.

\begin{table}[t]
    \centering
    \setlength{\abovecaptionskip}{0cm} 
    \setlength{\belowcaptionskip}{0.1cm}
    \caption{Summary of existing available real-world turbulence benchmarks.}
      \renewcommand{\arraystretch}{0.7}
      \setlength\tabcolsep{2.5pt}
      \begin{adjustbox}{width=0.9\columnwidth}
      \scriptsize
      \begin{tabular}{cccccccccccc>{\columncolor{gray!0}}c}
      \toprule[1pt]
       &Venue           &Hot-air       &Temperature  &Long-range    &Distance        & Scene Category     & Scene     &Avg Frames                 & Resolution  \\
      \midrule[1pt]
           & EFF \cite{hirsch2010efficient}       &\ding{51}    &-        &\ding{55}     &$<$1Km          &Building, Chimney  &2             &100                         & 240*240       \\
           & Heat Chamber \cite{mao2022single}     & \ding{51}    &-               &\ding{55}     &20m           &Image    &400            &100                        &440*440            \\
           & UG2+ TurbuText \cite{Textdataset}     & \ding{51}    &-               &\ding{55}     &20m           &Text    &100            &100                        &440*440            \\
           & CLEAR \cite{anantrasirichai2013atmospheric}      & \ding{51}  &$46^\circ$                &\ding{55}      &$<$2Km          &Building, Street     &3             &53                     &250*180            \\
           & OTIS \cite{gilles2017open}      & \ding{51}    &-              &\ding{55}     &$<$1Km        &Pattern, door       &21             &276                           &256*256 \\
           & TSR-WGAN \cite{jin2021neutralizing}      &\ding{51}    &$33^\circ$          &\ding{55}    &$<$3Km    &Street, Grassland        &21           &233                  &960*540      \\
           & Turbulence Text \cite{mao2022single}     & \ding{51}    &$30^\circ$                 &\ding{55}     &300m           &Text     &100            &100                        &440*440            \\
           & TurbRecon \cite{mao2020image}       &\ding{51}    &$30^\circ$             &\ding{51}     &4Km     &Building, Chimney          &4            &100                   &512*512             \\
           &\textbf{RLR-AT} &\ding{51}     &$-6^\circ \sim 40^\circ$       &\ding{51}     & \textbf{1Km-13Km}  &Building, Text $\ldots$ Car        &\textbf{1500}           &800   &1920*1080 \\
      \bottomrule[1pt]
      \end{tabular}
      \end{adjustbox}
      \label{tab_datasets}
\end{table}

\noindent\textbf{Atmospheric Turbulence Mitigation}. Lucky imaging is an intuitive way to mitigate the turbulence \cite{brandner2016lucky, law2006lucky,hua2020removing,chimitt2019rethinking,anantrasirichai2018atmospheric,boehrer2020turbulence,lau2019restoration}. Its key idea is to choose the lucky high-quality frames least affected by atmosphere from short-exposure imaging frames. Unfortunately, the lucky assumption does not hold any more for long-range anisoplanatic turbulence, where severe distortions persist across all frames \cite{roggemann1996imaging, fried1982anisoplanatism,mao2020image}. In recent years, the data-driven methods have been popular due to its end-to-end simplicity \cite{jaiswal2023physics, mao2022single, gao2019atmospheric, yasarla2021learning, mei2023ltt, li2021unsupervised, jiang2023nert, feng2022turbugan, rai2022removing, wang2023revelation,zhang2022imaging,zhang2024spatio}. Its main idea is to train on the paired clean-degraded synthetic data from turbulence simulators \cite{chimitt2022real, chimitt2020simulating, mao2021accelerating, schwartzman2017turbulence, hardie2017simulation}. The learning-based methods would achieve satisfactory results on simulated data while can not generalize well to the real turbulence due to the domain gap, especially for long-range turbulence with severe distortions. Considering the turbulence has clear physical procedure namely light refraction and diffraction, Zhu \emph{et al.} \cite{zhu2012removing} proposed a classical multi-stage restoration framework which gradually performed the distortion correction and deblurring. In this work, we follow this research line with clear physics foundations, and investigate how to cooperate the dynamic and static priors for better distortion correction.

\noindent\textbf{Turbulence Distortion Correction}. The main difficulty of the long-range turbulence lies in the severe distortions. Most existing methods can be classified into two categories:  registration-based methods \cite{shimizu2008super, zhu2012removing, caliskan2014atmospheric, xie2016removing, mao2020image, hua2020removing, fazlali2022atmospheric} and decomposition-based methods \cite{oreifej2012simultaneous, deshmukh2013fast, he2016atmospheric}. Most existing registration-based methods employ the zero-mean assumption prior of dynamic turbulence to construct a reference frame, and suppress distortion by aligning each input frame to the reference frame utilizing non-grid registration. For example, the most common way to obtain a reference frame is by directly averaging input frames \cite{hirsch2010efficient, zhu2012removing, anantrasirichai2013atmospheric, caliskan2014atmospheric, hardie2017block}. Mao \emph{et al.} \cite{mao2020image} further presented a novel space-time non-local averaging method to adaptively assign different weights for different frames, departing from uniform temporal averaging. However, these methods would suffer from blur, leading to inaccurate registration, especially for long-range turbulence.  In this work, we discover the pixel motion statistical prior of turbulence and propose a frequency-aware high-quality reference frame for better large-scale distortion registration.

The removal-based methods started from the matrix decomposition perspective, utilizing low-rank prior of static background to remove distortions. For example, Oreifej \textit{et al.} \cite{oreifej2012simultaneous} proposed a three-term matrix decomposition approach for simultaneously distortion removal and object detection. However, these matrix-based methods need to transform 3-D video into 2-D matrix, which would unexpectedly damage the spatio-temporal structure. In this work, we propose a subspace-based low-rank tensor refinement model to refine the registration error meanwhile well preserving the spatio-temporal details. We further integrate dynamic and static priors within a coarse-to-fine framework in a complementary manner to better handle long-range turbulence with severe distortions.

\begin{figure*}[t]
  \captionsetup{skip=1.5mm} 
  \centering
  \includegraphics[width=0.85\linewidth]{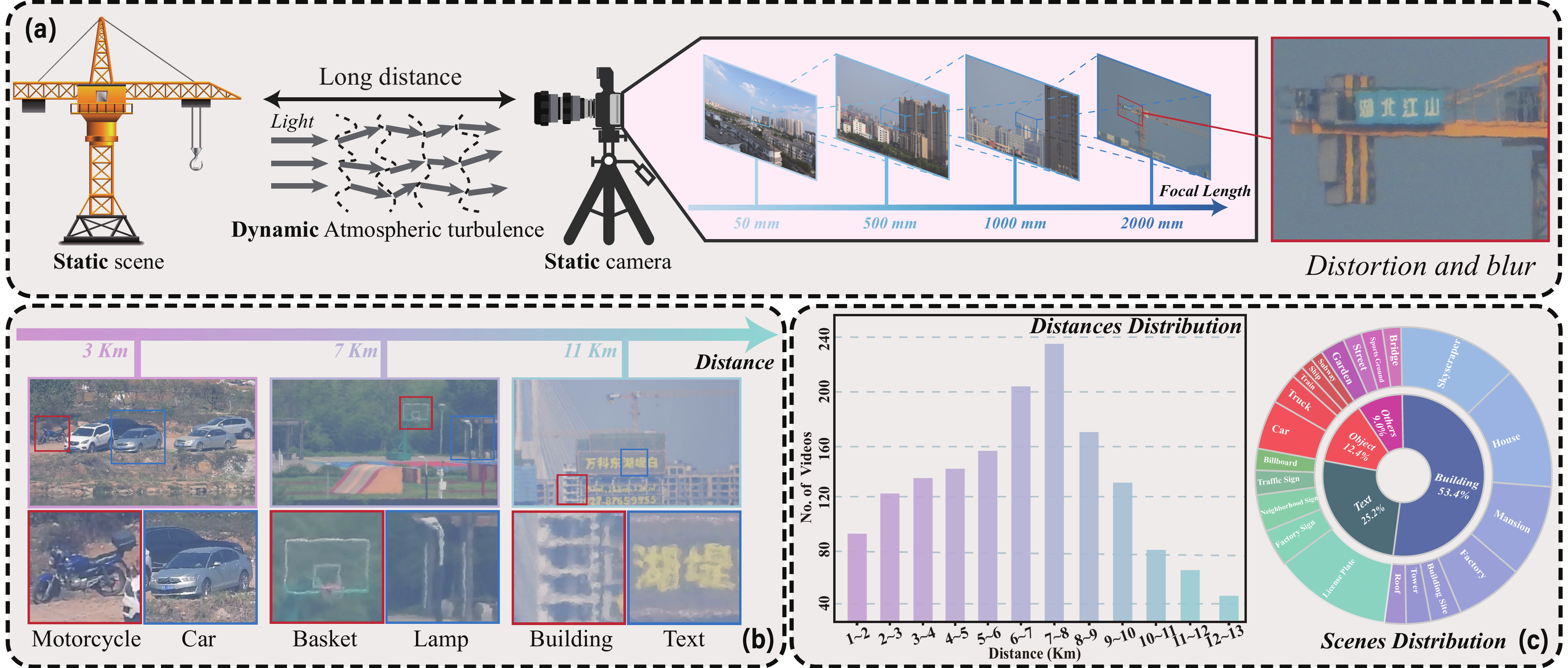}
  \caption{Illustration of the proposed dataset RLR-AT. (a) Long-range imaging with larger focal length lens through turbulence. (b) Typical turbulence with diverse long-distance conditions. (c) Statistics of distance and scene of the proposed benchmark.}
  \label{datasets}
\end{figure*}
\section{Large-scale Real Long-range Turbulence Benchmark}

Owing to the challenges in gathering long-range turbulence, current datasets mainly focus on  hot-air turbulence, overlooking  long-range turbulence with severe distortions. To fill this gap, we construct a large-scale long-range turbulence dataset for verification and analysis, named as RLR-AT. Note that RLR-AT also includes videos of \textit{dynamic scenes} and \textit{turbulence coupled with haze}, which can be used to study turbulence in dynamic scenes and multi-degradation restoration.

\noindent\textbf{Benchmark Collection.}
In this work, we collect the long-range turbulence sequences by a telephoto camera (Nikon Coolpix P1000) with \textit{equivalent} 3000mm lens focal length, sampled in 30 fps. The data collection process is illustrated in Fig. \ref{datasets}(a). Firstly, we stabilize the camera on a tripod to capture distant static scenes. Subsequently, we adjust the focal length until we discern the emergence of geometric distortion and blurring induced by non-uniform indices of refraction  associated with long-range turbulence. For each sequence, we record approximate 35 seconds and extract the intermediate steady 30 seconds into our dataset.

\noindent\textbf{Benchmark Statistics.}
Table \ref{tab_datasets} presents the detailed statistical comparison between our proposed RLR-AT and existing turbulence benchmarks. Overall, our dataset contains 1500 sequences, each of which consists of approximately 800 frames, collected from diverse cities. Notably, our dataset comprehensively covers long-range turbulence across distances ranging from 1 km to 13 km. Moreover, over 19 typical scenes are captured, including the street, sports ground, factory, car and billboard, etc, offering a comprehensive range for long-range surveillance scenarios. To visualize the distribution of distances and scene categories, a bar chart and a sunburst chart are illustrated in Fig. \ref{datasets}(c).

\noindent\textbf{Turbulence with Longer-distances and Higher-diversity.}
The key difference between RLR-AT and other datasets is that RLR-AT covers turbulence distortions with longer-distances and higher-diversity. Most existing public datasets are mainly composed of hot-air turbulence, and TurbRecon \cite{mao2020image} stands out by capturing two turbulence sequences at a distance of 4 Km. In comparison, RLR-AT contains long-range turbulence images captured from longer and richer distances (1-13Km). In Fig. \ref{datasets}(b), we visualize some long-range turbulence images with increasing distances in RLR-AT. It can be observed that with increasing distance, turbulence degradation level is higher, leading to more severe distortions in the images. The bar chart in Fig. \ref{datasets}(c) displays the number of turbulence sequences in our dataset at each distance, ranging from 1 km to 13 km, which further highlights the diversity of distances within our RLR-AT.

\noindent\textbf{Scenes with Larger-scale and Greater-variety.}
We concern not only the distances diversity of long-range turbulence but also the variety and scale of scenes. In Table \ref{tab_datasets}, we can observe that the existing real-world turbulence benchmarks are still limited in terms of scene amount and diversity. Most datasets focus on relatively common scenes, such as  building, street or grassland, and UG2+ TurbuText \cite{Textdataset}  and Turbulence Text \cite{mao2022single} both collect hot-air turbulence specifically for text scenes. In comparison, RLR-AT contains 1500 long-range turbulence sequences across 19 various categories. Figure \ref{datasets}(b) shows some typical long-range turbulence images in various scenes, such as car, motorcycle, building, text, offering a comprehensive range for long-range surveillance scenarios. The sunburst chart in Fig. \ref{datasets}(c) illustrates the distribution of scene categories in RLR-AT, further showcasing the diversity of scenes within our dataset.

\begin{figure*}[t]
  \centering
  \captionsetup{skip=1.3mm} 
     \includegraphics[width=0.85\linewidth]{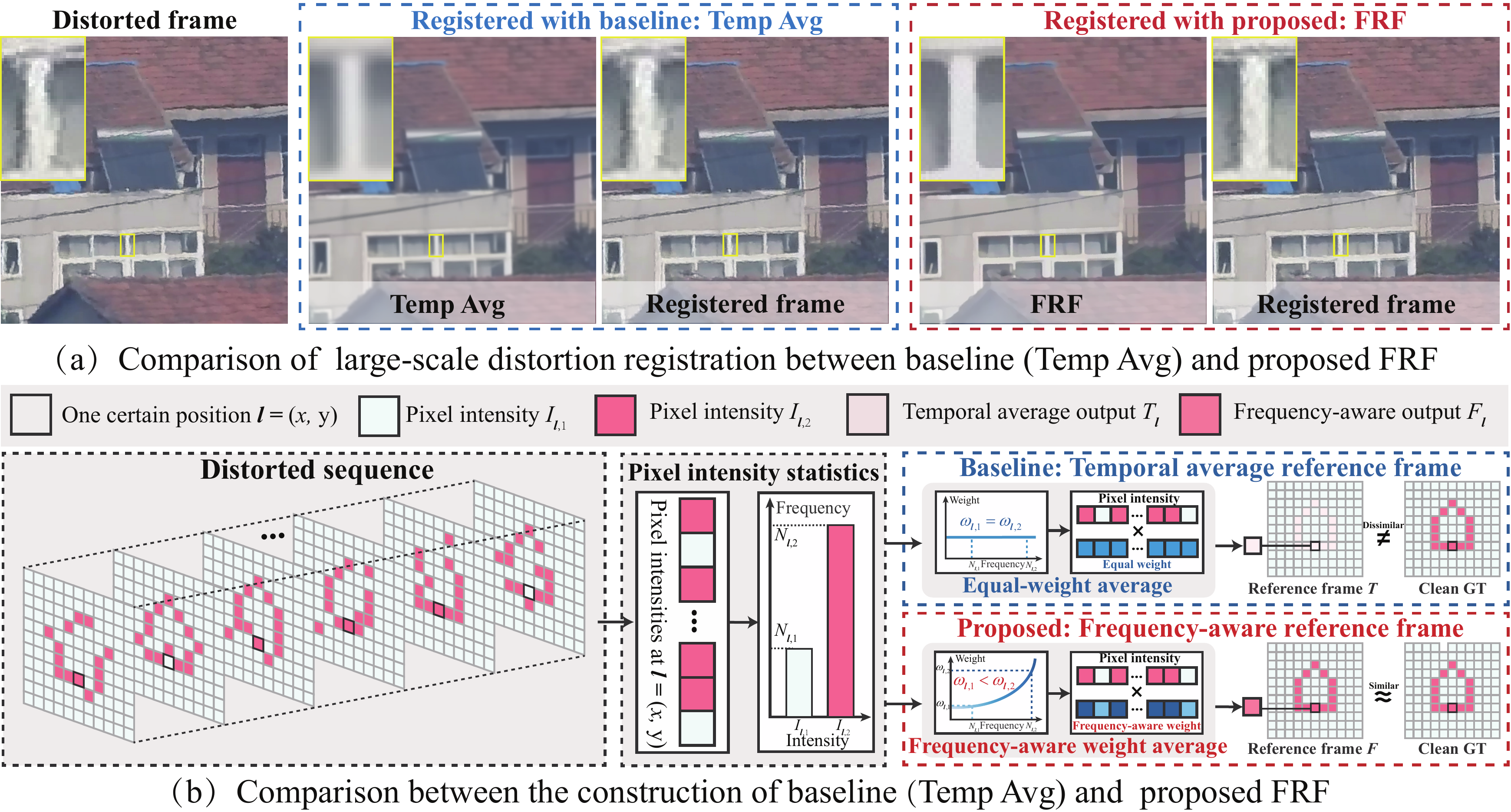}
  \caption{Comparison between Temp Avg and proposed FRF. (a) Comparison of registration performance between Temp Avg and FRF. Temp Avg fails to achieve precise registration since it suffers from blur, while FRF achieves more accurate registration. (b) Comparison between the construction of Temp Avg and FRF. Temp Avg assigns equal weight to all intensities at a certain position. However, the output is dissimilar with
  original GT, since the less frequently occurring intensities make a negative contribution. In contrast, we argue that the higher
  the frequency, the greater the weight, as the most frequently occurring pixel at the position is closer to the original GT.}
  \label{referenceframe}
\end{figure*}

\section{A Coarse-to-fine Framework for Long-range Turbulence}\label{sec4}
In this work, we propose a coarse-to-fine framework that cooperates dynamic turbulence prior and static background priors (CDSP) to handle long-range turbulence with severe distortions. On the one hand, we discover the pixel motion statistical prior of turbulence and propose a frequency-aware reference frame for better large-scale distortion registration, which greatly reduces the burden of refinement (Section \ref{sec4.1}). Then we align the distorted frames to the proposed reference frame utilizing registration approach based on optical flow \cite{liu2009beyond}. On the other hand, we
take advantage of the static prior of background and propose a subspace-based low-rank tensor refinement model to  refine the registration errors unavoidably left by registration while well preserving details (Section \ref{sec4.2}). The dynamic and  static priors complement to each other, facilitating us to better eliminate the
severe distortions. Finally, we employ a simple data-driven network to further remove the residual blur, and the generation of paired deblurring data is based on the proposed distortion correction framework. The details of blur removal are provided in the supplementary material.

\subsection{Frequency-aware Reference Frame Construction}\label{sec4.1}
Most previous methods \cite{hirsch2010efficient,caliskan2014atmospheric,hardie2017block,zhu2012removing} employ temporal averaging (Temp Avg) to construct a reference frame by naively applying zero-mean assumption of turbulence. However, Temp Avg often suffers from blur, leading to imprecise registration when handling severe distortions, as shown in Fig. \ref{referenceframe}(a). In contrast, FRF achieves more accurate registration due to its superior quality. In Fig. \ref{referenceframe}(b), Temp Avg assumes that all pixel intensities occurring at a certain position have equal weight. However, the output is dissimilar
with original GT, as the non-original pixel intensities make a negative contribution to the output at the certain position. In this work, we propose a frequency-aware method to construct a reliable reference frame based on the pixel motion statistical prior of turbulence.

\begin{figure*}[t]
  \captionsetup{skip=0.8mm} 
  \centering
     \includegraphics[width=0.90\linewidth]{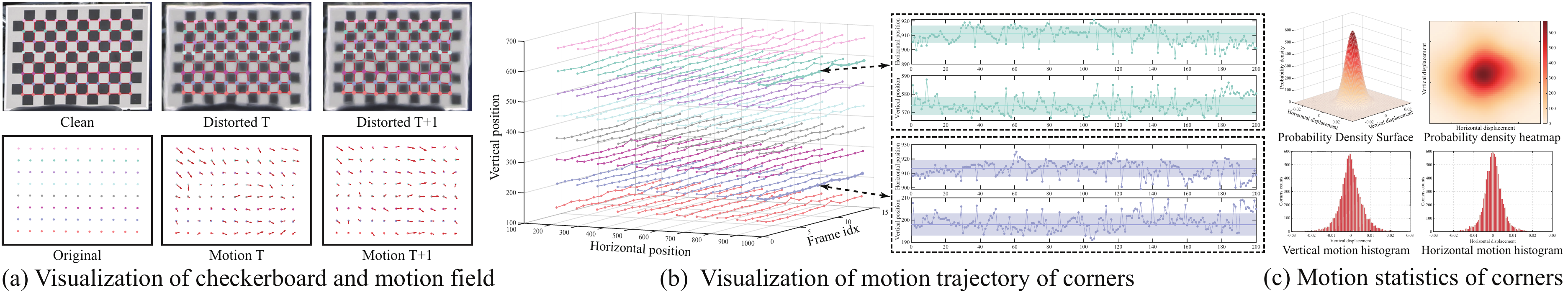}
  \caption{Analysis of pixel motion statistical priors using checkerboard corners.(a) visualizes the clean and distorted checkerboard and the motion field of corners. (b) shows the motion trajectory of corners along the temporal axis, with an extended display of two randomly selected corners across 200 frames. (c) performs a statistic of corner motions and the motions of corners approximately conform to a zero-mean gaussian distribution, indicating that corners most frequently occur in their original positions.}
  \label{dynamic_prior}
\end{figure*}

\noindent\textbf{Pixel Displacement Statistical Prior of Turbulence.}
To explore the pixel displacement statistical prior of turbulence, we conduct an analysis experiment utilizing the corners of checkerboard in Fig. \ref{dynamic_prior}. We take checkerboard turbulence images as the experimental datasets due to the ability to approximate corner motion as pixel motion, and the mature nature of checkerboard corner detection techniques. Then
we apply a corner detector \cite{geiger2012automatic} on the collected datasets to detect the shifted corners. Note that we conduct extensive analysis experiments on long-range turbulence across various distances and scenes ( \textit{e.g.} Building with wall corners). Please refer to supplementary material for details.

In Fig. \ref{dynamic_prior}(a), we show the clean and distorted checkerboard and corresponding motion field of corners. It is evident that the corners of distorted frames exhibit noticeable motion, and the motion of corners varies across different frames. To better explore the motion of corners in the temporal dimension, we further visualize the motion trajectory of the corners along the temporal axis in Fig. \ref{dynamic_prior}(b). The motion trajectories of each corner are distinct. We randomly select two corners and present their horizontal and vertical positions on the right. It is observed that although the displacements of the two corners differ, a commonality is that they consistently occur relatively close to their original positions, resembling a statistical rule. To further explore the statistics of pixel motion, we normalize the positions of all corners across 200 frames and perform statistics analysis on their displacements in Fig. \ref{dynamic_prior}(c). The statistical results show that the motions approximately follow a zero-mean gaussian distribution, indicating that pixels most frequently occur in their original positions. This insight implies that \textit{the pixel occurring most frequently at one certain position is most likely the original GT}, inspiring us
to propose a frequency-aware reference frame. 

\noindent\textbf{Frequency-aware Reference Frame.} 
Given the distorted sequence, suppose there are $K$ different pixel intensities occurring at a certain position $\bm{\mathit{l}} = (x,y)$ along the temporal dimension. Let $I_{\bm{\mathit{l}},k}$ represent a pixel intensity occurring at this position in the distorted sequence and $k \in [1,K]$. We first count the frequency of each pixel intensity occurring at the position along the temporal dimension: $N_{\bm{\mathit{l}},k}= count(I_{\bm{\mathit{l}},k})$.
Previous temporal average reference frame assigns a weight of one to each pixel intensity at the certain position, hence, the output $T_{\bm{\mathit{l}}}$ is obtained by summing all pixel intensities and dividing by the number of frames that equals the sum of the frequency of pixel intensities:
\begin{equation}
  \setlength{\abovedisplayskip}{0pt}
  \setlength{\belowdisplayskip}{0pt}
  \resizebox{0.42\linewidth}{!}{$
  \begin{aligned}
    T_{\bm{\mathit{l}}}  = \Big( {\sum\limits_{k=1}^{K} N_{\bm{\mathit{l}},k}\times I_{\bm{\mathit{l}},k}  } \Big)/ \Big({\sum\limits_{k=1}^{K} N_{\bm{\mathit{l}},k}}\Big),
  \end{aligned}
  $}
  \label{eq:avgref}
\end{equation}
Different from temporal averaging,  we argue that the weight of pixel intensities is positively correlated with their frequency as shown in Fig. \ref{referenceframe}(b). Consequently, we construct a frequency-aware weight for each pixel intensity:
 \begin{equation}
  \setlength{\abovedisplayskip}{0pt}
  \setlength{\belowdisplayskip}{0pt}
  \begin{aligned}
  \omega_{\bm{\mathit{l}},k} = \mathrm{e}^{\sigma \times N_{\bm{\mathit{l}},k}},
  \end{aligned}
  \label{eq:weight}
\end{equation}
which is a function of the frequency $N_{\bm{\mathit{l}},k}$, and $\sigma$ is a hyper-parameter controlling the growth rate of the weight. Next, the pixel value of the reference frame at the certain position $F_{\bm{\mathit{l}}}$ is constructed via weighted averaging based on frequency:
\begin{equation}
  \setlength{\abovedisplayskip}{0pt}
  \setlength{\belowdisplayskip}{0pt}
  \resizebox{0.55\linewidth}{!}{$
  \begin{aligned}
  F_{\bm{\mathit{l}}}  = \Big({\sum\limits_{k=1}^{K} N_{\bm{\mathit{l}},k}\times I_{\bm{\mathit{l}},k} \times \omega_{\bm{\mathit{l}},k}  } \Big) / \Big({\sum\limits_{k=1}^{K} N_{\bm{\mathit{l}},k}} \times  \omega_{\bm{\mathit{l}},k} \Big).
  \end{aligned}
  $}
  \label{eq:ref}
\end{equation}
\noindent\textbf{Relationship between FRF and Temp Avg.}  We further discuss the relationship between Temp Avg and FRF, which is established through the parameter $\sigma$. The $\sigma$ in Eq. (\ref{eq:weight}) decides the sensitivity of weight function to frequency.  When $\sigma = 0$, the weight is one for all intensities, and the Eq. (\ref{eq:ref}) can be simplified as:
\begin{equation}
  \setlength{\abovedisplayskip}{0pt}
  \setlength{\belowdisplayskip}{0pt}
  \resizebox{0.4\linewidth}{!}{$
  \begin{aligned}
    F_{\bm{\mathit{l}}}  = \Big( {\sum\limits_{k=1}^{K} N_{\bm{\mathit{l}},k}\times D_{\bm{\mathit{l}},k}  } \Big)/ \Big({\sum\limits_{k=1}^{K} N_{\bm{\mathit{l}},k}}\Big).
  \end{aligned}
  $}
  \label{eq:Favgref}
\end{equation}
which is the same as Eq. (\ref{eq:avgref}), illustrating that average reference frame is a special case of proposed FRF when $\sigma = 0$. The reference frame constructed with $\sigma = 0$ (Temp Avg) is shown in Fig. \ref{referenceframe}(a), it can be observed that the result suffers from severe blur. On the contrary, when $\sigma \neq 0$, the higher the frequency of pixel intensity, the greater the weight, resulting in a output more similar to the original pixel. The result with $\sigma \neq 0$ (FRF) in Fig. \ref{referenceframe}(a) possesses superior visual 
quality compared to Temp Avg, which is beneficial for severe distortion registration.

\begin{figure}[t]
  \centering
  \captionsetup{skip=0.8mm} 
     \includegraphics[width=0.90\linewidth]{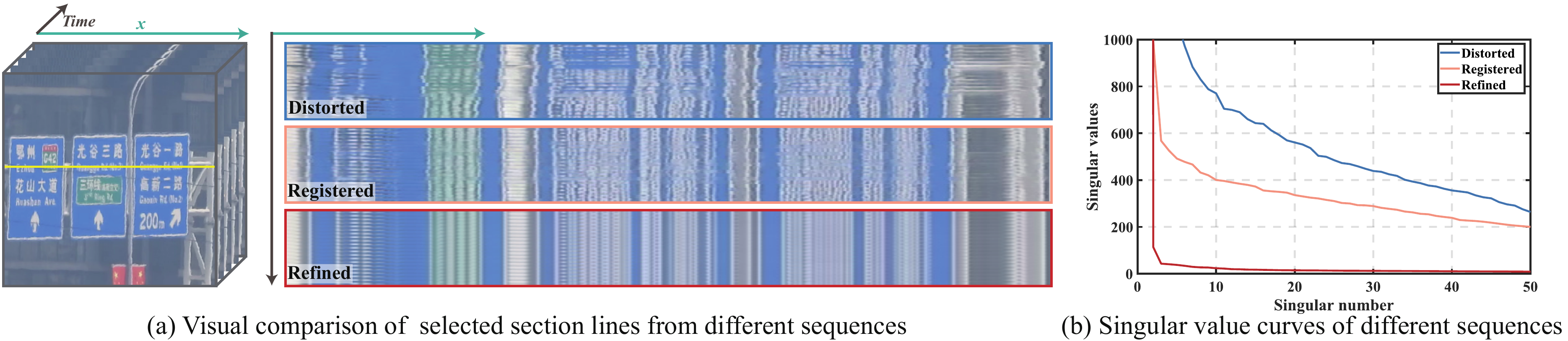}
  \caption{Low-rank property analysis of different sequences. (a) Visualization of selected section lines. (b) Singular value curves of corresponding sequences.}
  \label{Static}
\end{figure}

\subsection{Low-rank Tensor Distortion Refinement}\label{sec4.2}

\noindent\textbf{Low-rank Prior of Static Background.}
Due to the severe distortions in long-range turbulence, achieving perfect pixel-level registration is impossible. Consequently, registration errors are unavoidable in the registered sequences. Considering the static nature of scene, we aim to utilize low-rank prior of static background to refine registration errors while preserving details.
We utilize the section lines and singular values to analyze the low-rank property of the distorted, registered and refined sequence in Fig. \ref{Static}. In Fig. \ref{Static}(a),  we randomly select 1D section lines from each sequence. It is observed that the registered section lines contain sparse noise, while the refined section lines exhibit smoothness along the temporal dimension. Figure \ref{Static}(b) shows the curves of singular value, revealing that the refined sequence manifests the strongest low-rank property.

\noindent\textbf{Subspace-based Low-rank Tensor Refinement Model.}
Previous methods directly utilized matrix decomposition to remove turbulence \cite{oreifej2012simultaneous,he2016atmospheric}, which need to transform 3-D video into 2-D matrix, damaging the spatial-temporal structure. In this work, we propose a subspace-based low-rank tensor refinement model (SLRTR) to rectify the misalignments while preserving details. To our knowledge, we are the first to introduce tensor model into turbulence removal.
Given the registered sequence $\boldsymbol{\mathcal{R}}\in{{\mathbb{R}}^{h \times w \times t}}$, where $h$, $w$ and $t$ respectively denote the image height, width, and the number of frames. The key challenge lies in effectively reducing registration error while preserving the spatio-temporal details. A registered sequence can be described as the following formula:
\begin{equation}
  \setlength{\abovedisplayskip}{0pt}
  \setlength{\belowdisplayskip}{0pt}
  {\boldsymbol{\mathcal{R}}}  = {\boldsymbol{\mathcal{B}}} + {\boldsymbol{\mathcal{E}}} + {\boldsymbol{\mathcal{N}}},
  \label{eq:syn3}
  \end{equation}
where ${\boldsymbol{\mathcal{B}}}\in{{\mathbb{R}}^{h \times w \times t}}$ represents the refined sequence, ${\boldsymbol{\mathcal{E}}} \in{{\mathbb{R}}^{h \times w \times t}}$ is the registration error,  ${\boldsymbol{\mathcal{N}}}\in{{\mathbb{R}}^{h \times w \times t}}$ denotes the random noise. In this work, we formulate the refinement as an inverse problem utilizing the \textit{maximum-a-posterior}, as follows:
  \begin{equation}
  \setlength{\abovedisplayskip}{0pt}
  \setlength{\belowdisplayskip}{0pt}
  \resizebox{0.51\linewidth}{!}{
   $\mathop {\min }\limits_{{\boldsymbol{\mathcal{B}}},{\boldsymbol{\mathcal{E}}}} \frac{1}{2}||{\boldsymbol{\mathcal{B}}} + {\boldsymbol{\mathcal{E}}} -  {\boldsymbol{\mathcal{R}}} || _F^2 +\alpha \Phi_b ({\boldsymbol{\mathcal{B}}}) +\beta \Phi_e ({\boldsymbol{\mathcal{E}}}),$}
   \label{eq:syn4}
  \end{equation}
where $\Phi_b$ and $\Phi_e$ represent the prior knowledge for the background and error, respectively, $\alpha$ and $\beta$ are the corresponding hyper-parameters. For static scene turbulence videos, on the one hand, the refined sequence $ {\boldsymbol{\mathcal{B}}} $ exhibits global low-rank property along the temporal dimension, with an ideal rank of one. On the other hand, it also has significant non-local low-rank property along the spatial dimension, due to the self-similarity widely employed in image restoration \cite{dabov2007image}. Hence, we effectively exploit a joint global-nonlocal prior across both spatial and temporal dimensions to enhance the representation of the static background $ {\boldsymbol{\mathcal{B}}} $:
  \begin{equation}
  \setlength{\abovedisplayskip}{2pt}
  \setlength{\belowdisplayskip}{0pt}
  \begin{aligned}
  &\resizebox{0.55\hsize}{!}{$\Phi_b ({\boldsymbol{\mathcal{B}}}) =  \alpha \mathop \sum \limits_i \left(\frac{1}{{\lambda _i^2}}||{{\boldsymbol{\mathcal{S}}}_i}{\boldsymbol{\mathcal{B}}}{ \times _3}O_i - {{\boldsymbol{\mathcal{G}}}_i}|| _F^2 +||{\boldsymbol{\mathcal{G}}}_i||_{tnn}\right),$}
  \end{aligned}
  \label{eq:syn4-1}
  \end{equation}
  where ${{\boldsymbol{\mathcal{S}}}_i}{\boldsymbol{\mathcal{B}}} \in {{\mathbb{R}}^{{p^2} \times n \times t}}$ is the constructed 3-D tensor via the non-local clustering of a sub-cubic ${u_i} \in {{\mathbb{R}}^{p \times p \times t}}$ \cite{chang2017hyper}, $p$ and $n$ are the spatial size and number of the sub-cubic respectively, $O_i \in {{\mathbb{R}}^{d \times t}}(d \ll t)$ is an orthogonal subspace projection matrix used to capture the temporal low-rank property, $\times _3$ is the tensor product along the temporal dimension \cite{kolda2009tensor}, ${{{\boldsymbol{\mathcal{G}}}_i}}$ represents the low-rank approximation variable, $||\bullet||_{tnn}$ means the tensor nuclear norm for simplicity \cite{chang2017hyper}, $\lambda _i$ is the regularization parameter. As for the error ${\boldsymbol{\mathcal{E}}}$, we formulate it as the sparse error \cite{wright2009robust} via the $L_1$ sparsity. Thus, the Eq. (\ref{eq:syn4}) can be expressed as:
  \begin{equation}
  \setlength{\abovedisplayskip}{2pt}
  \setlength{\belowdisplayskip}{2pt}
  \begin{aligned}
  &\resizebox{0.55\hsize}{!}{$\left\{ {\hat {\boldsymbol{\mathcal{B}}}, \hat {\boldsymbol{\mathcal{E}}}, {{\hat {\boldsymbol{\mathcal{G}}}}_i}, \hat O_i} \right\} = \arg \mathop {\min }\limits_{{\boldsymbol{\mathcal{B}}},{\boldsymbol{\mathcal{E}}},{{\boldsymbol{\mathcal{G}}}_i},O_i} \frac{1}{2}|| {\boldsymbol{\mathcal{B}}} + {\boldsymbol{\mathcal{E}}} -  {\boldsymbol{\mathcal{R}}} || _F^2$}\\
  &\resizebox{0.55\hsize}{!}{$+ \beta ||{\boldsymbol{\mathcal{E}}}{|| _1}  + \alpha \mathop \sum \limits_i \left( {\frac{1}{{\lambda _i^2}}||{{\boldsymbol{\mathcal{S}}}_i}{\boldsymbol{\mathcal{B}}}{ \times _3}O_i - {{\boldsymbol{\mathcal{G}}}_i}||_F^2 + ||{\boldsymbol{\mathcal{G}}}_i||_{tnn}} \right).$}
  \end{aligned}
  \label{eq:syn4-2}
  \end{equation}
  To solve $ {\boldsymbol{\mathcal{B}}}, {\boldsymbol{\mathcal{E}}}, {\boldsymbol{\mathcal{G}}}_i, O_i$, we adopt the alternating minimization scheme \cite{lin2011linearized} to solve the Eq. (\ref{eq:syn4-2}) for each variable.  Please refer to the appendix for the solution.


\section{Experimental Results}

\begin{figure*}[t]
  \centering
  \captionsetup{skip=0.7mm} 
     \includegraphics[width=0.90\linewidth]{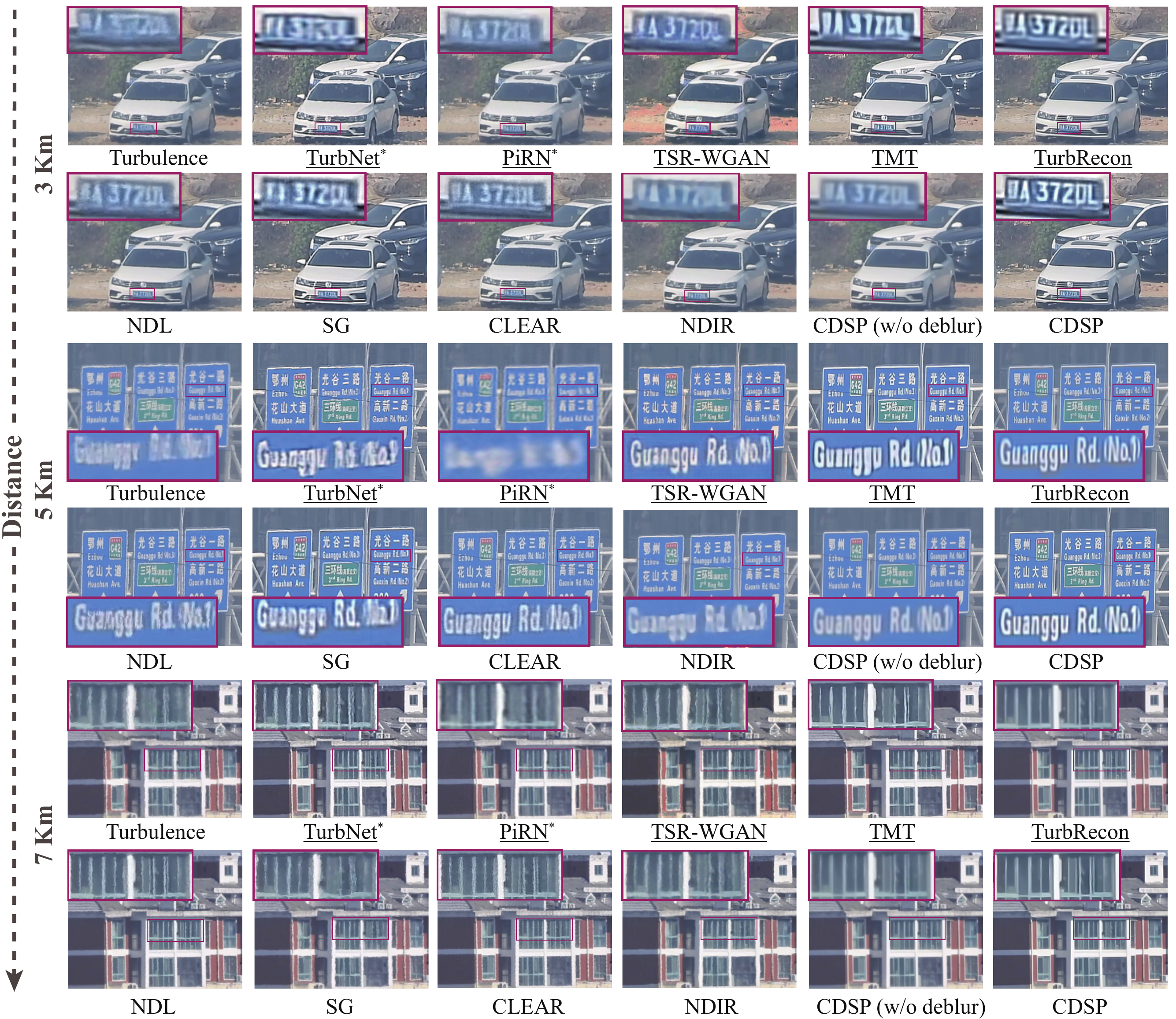}
  \caption{Visual comparisons of long-range turbulence mitigation at various distances on  RLR-AT. Figures on the above row are results of static-scene independent methods (marked by \underline{method}), and $^*$ denotes single-frame based method. Figures on the below row are results of static-scene dependent methods.  }
  \label{exp_real}
\end{figure*}

\begin{table}[t]
  \centering
  \setlength{\abovecaptionskip}{0cm} 
  \setlength{\belowcaptionskip}{0.1cm}
  \setlength{\arrayrulewidth}{1pt}
  \caption{ Quantitative comparison with other methods on synthetic turbulence dataset at different distances. $^*$ denotes single-frame based approach. \textcolor[rgb]{0.78,0.14,0.14}{\textbf{Red}} text indicates the best performance, \textcolor[rgb]{0.15,0.47,0.71}{\textbf{blue}} text indicates the second-best performance. \textbf{\textit{$\Delta$}} denotes the exact superiority of proposed CDSP over the second-best method. As the distance increases, our method outperforms the SOTA approach even more significantly.}
  \captionsetup{position=bottom} 
  \renewcommand{\arraystretch}{1}

  \setlength\tabcolsep{1.3pt}
  \begin{adjustbox}{width=0.95\columnwidth}
  \normalsize
  \centering
  \begin{tabular}{c|c|ccccc|ccccc|c>{\columncolor{gray!0}}c}
  \hline
  \multirow{2}*{\textbf{{Distance}}}&\multirow{2}*{\textbf{Metric}}&\multicolumn{5}{c|}{\textbf{{Static-scene Independent Methods}}}&\multicolumn{5}{c|}{\textbf{{Static-scene Dependent Methods}}} &\multirow{2}*{\textbf{\textit{$\Delta$}}}\\ \cline{3-12}
  & &TurbNet$^*$ &PiRN$^*$     & TSR-WGAN   &TMT &TurbRecon	&NDL   &CLEAR       &SG  & NDIR &CDSP \\ 
  \specialrule{1pt}{0pt}{0pt} 
  \rowcolor{gray!25}
  \cellcolor{white}\multirow{2}*{\textbf{\textit{2 Km}}}&PSNR$\uparrow$    &23.58 &25.93 &23.98  &26.54 &\textcolor[rgb]{0.15,0.47,0.71}{\textbf{27.27}}     &24.93  &26.19 &24.53  &24.29  &\textcolor[rgb]{0.78,0.14,0.14}{\textbf{27.94}} &\textbf{0.67}\\ 
  &SSIM$\uparrow$  &0.8155 &0.8675 &0.8215  &0.8887   &\textcolor[rgb]{0.15,0.47,0.71}{\textbf{0.8995}}&0.8459 &0.8815 &0.8545 &0.8147 &\textcolor[rgb]{0.78,0.14,0.14}{\textbf{0.9181}} &\textbf{0.0186}\\
  \hline
  \rowcolor{gray!25}
  \cellcolor{white}\multirow{2}*{\textbf{\textit{4 Km}}}&PSNR$\uparrow$    &21.97 &24.43 &22.90  &24.82&\textcolor[rgb]{0.15,0.47,0.71}{\textbf{26.05}}     &24.12  &24.79 &23.31 &23.23 &\textcolor[rgb]{0.78,0.14,0.14}{\textbf{26.91}}&\textbf{0.86}\\ 
  &SSIM$\uparrow$  &0.7491 &0.8168 &0.7813  &0.8556   &\textcolor[rgb]{0.15,0.47,0.71}{\textbf{0.8615}}&0.8111 &0.8382 &0.8026 &0.7796 &\textcolor[rgb]{0.78,0.14,0.14}{\textbf{0.8893}}&\textbf{0.0278}\\ 
  \hline
  \rowcolor{gray!25}
  \cellcolor{white}\multirow{2}*{\textbf{\textit{6 Km}}}&PSNR$\uparrow$    &21.22 &23.42 &21.89  &23.73 &\textcolor[rgb]{0.15,0.47,0.71}{\textbf{24.93}}     &23.37  &23.58 &22.50 &22.59  &\textcolor[rgb]{0.78,0.14,0.14}{\textbf{25.74}} &\textbf{0.81}\\ 
  &SSIM$\uparrow$  &0.7132 &0.7790 &0.7394 &0.8265  &\textcolor[rgb]{0.15,0.47,0.71}{\textbf{0.8311}}&0.7834 &0.7925 &0.7643 &0.7561 &\textcolor[rgb]{0.78,0.14,0.14}{\textbf{0.8594}} &\textbf{0.0283}\\ 
  \hline
  \rowcolor{gray!25}
  \cellcolor{white}\multirow{2}*{\textbf{\textit{8 Km}}}&PSNR$\uparrow$    &20.69 &22.74 &21.31  &23.01&\textcolor[rgb]{0.15,0.47,0.71}{\textbf{23.95}}     &22.76  &22.62 &21.89  &22.19  &\textcolor[rgb]{0.78,0.14,0.14}{\textbf{24.84}} &\textbf{0.89}\\ 
  &SSIM$\uparrow$  &0.6864 &0.7518 &0.7135  &0.8012    &\textcolor[rgb]{0.15,0.47,0.71}{\textbf{0.8024}}&0.7592 &0.7523 &0.7335 &0.7402 &\textcolor[rgb]{0.78,0.14,0.14}{\textbf{0.8319}}  &\textbf{0.0295}\\ 
  \hline
  \end{tabular}
  \end{adjustbox}
  \label{comparison}
\end{table}

\begin{figure}[t]
  \centering
  \begin{adjustbox}{width=0.95\columnwidth}
	\begin{minipage}[t]{0.63\linewidth}
    \captionsetup{skip=1.1mm} 
		\includegraphics[width=1\linewidth]{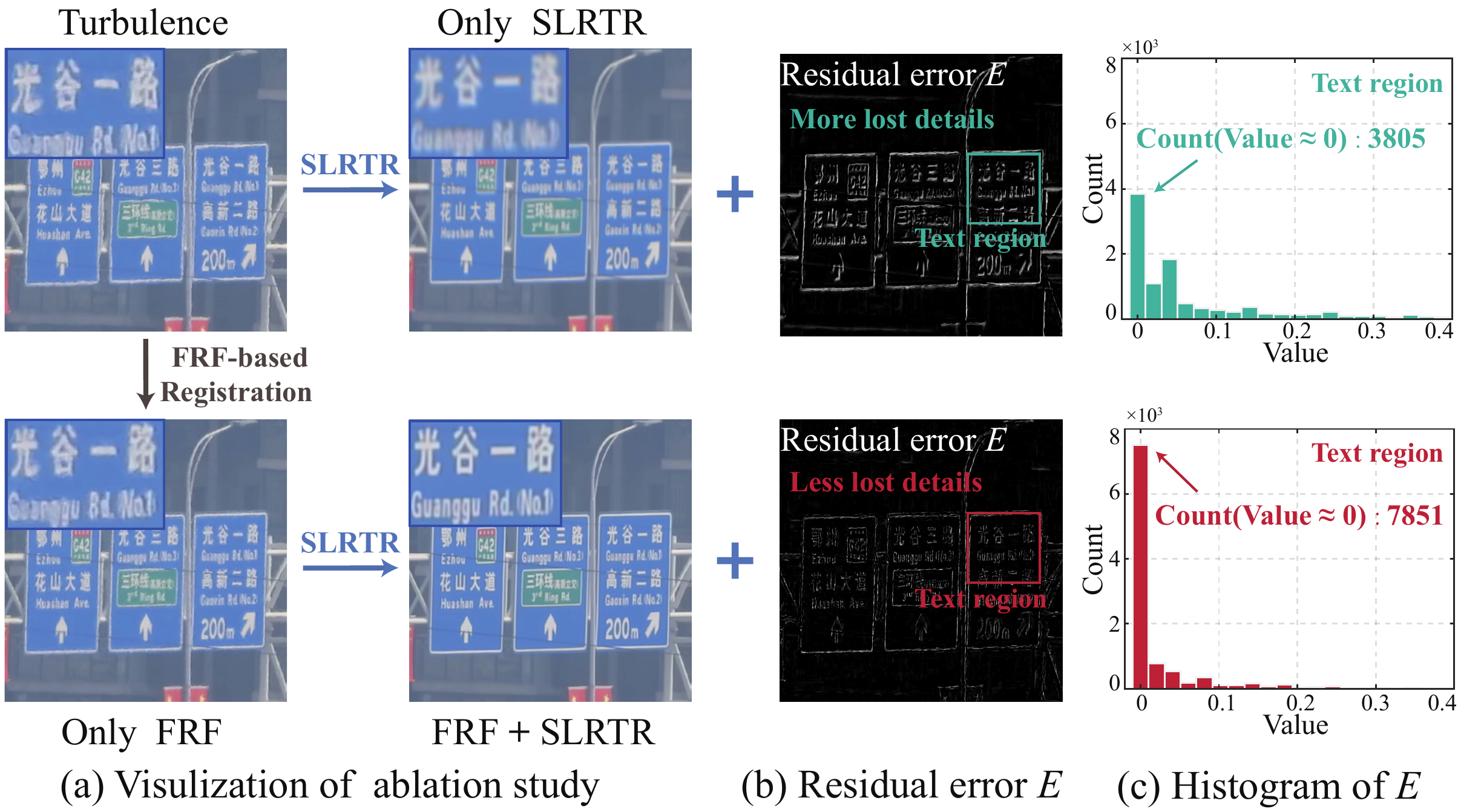}
		\caption{Effectiveness of FRF and SLRTR. (a) Visual comparison of results w/o FRF-based registration, w/o SLRTR and w/ both. (b) Residual error from SLRTR decomposition. (c) Histogram of residual error.  }
		\label{EFF_fig}
	\end{minipage}
  \hfill
	\begin{minipage}[t]{0.31\linewidth}
    \captionsetup{skip=1.1mm} 
		\includegraphics[width=1\linewidth]{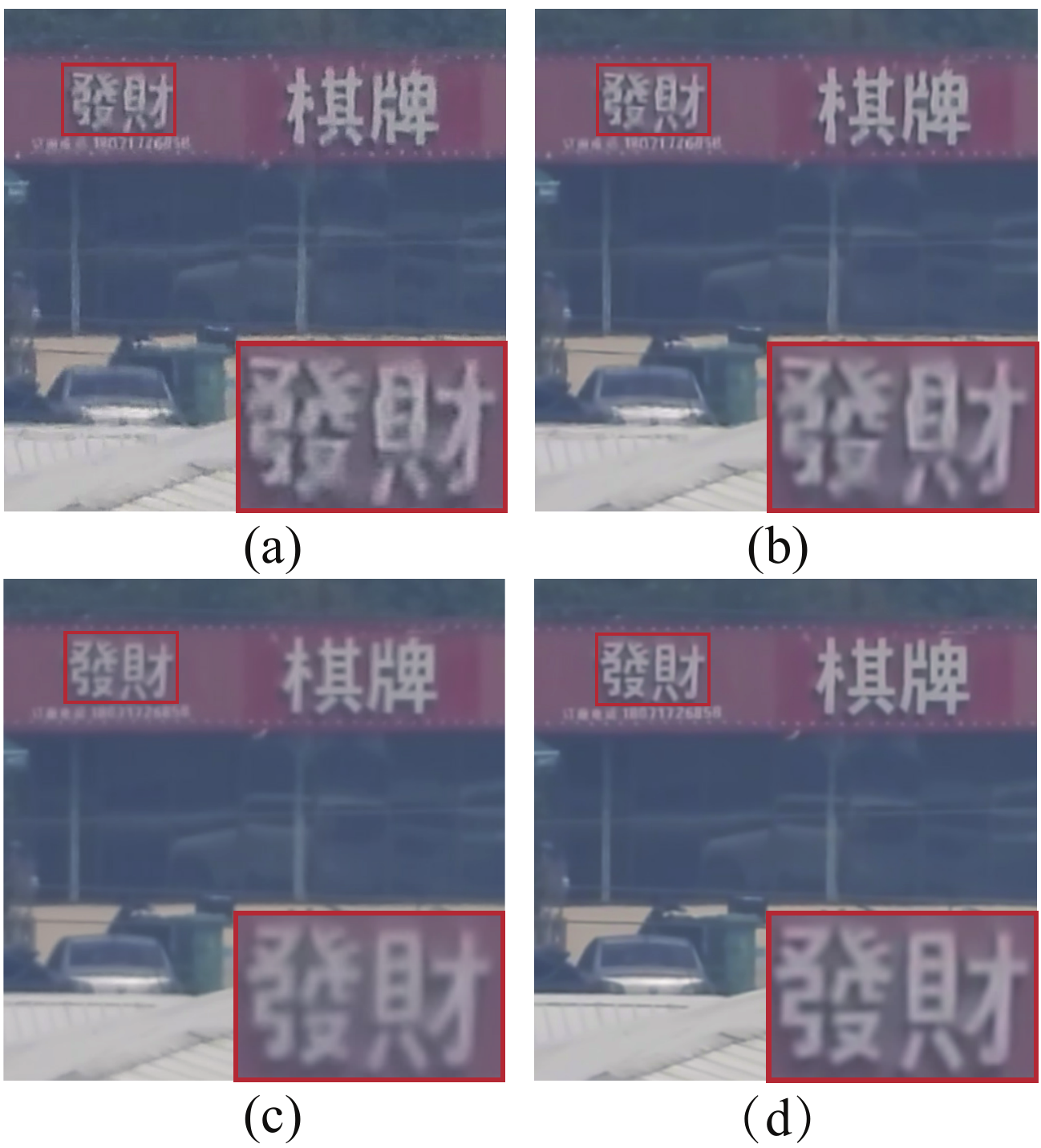}
		\caption{Ablation study on SLRTR. (a) Input. (b) w/o subspace. (c) w/o self-similarity. (d) w/ both.  }
		\label{EFF_SLRTR}
	\end{minipage}
  
\end{adjustbox}
\end{figure}

\subsection{Datasets and Experimental Settings}\label{sec5.1}
\noindent\textbf{Datasets.} 
We conduct the experiments on various datasets, including a synthetic dataset, the proposed dataset RLR-AT and real hot-air turbulence dataset TurbuText \cite{Textdataset}. Synthetic turbulence is simulated with varying distances on the ADE20K \cite{zhou2017scene} employing the turbulence simulator P2S \cite{mao2021accelerating}. Further details of the simulator protocol are provided in the supplementary material.

\noindent\textbf{Comparison Methods.}   
We compare CDSP with (1) conventional turbulence removal methods:  TurbRecon \cite{mao2020image}, SG \cite{lou2013video}, CLEAR \cite{anantrasirichai2013atmospheric} and NDL \cite{zhu2012removing}; (2) deep learning based methods: TurbNet \cite{mao2022single}, TSR-WGAN \cite{jin2021neutralizing}, PiRN \cite{jaiswal2023physics}, NDIR \cite{li2021unsupervised} and TMT \cite{zhang2022imaging}. For a fair comparison, considering that the experimental datasets consist of static scenes, we categorize the methods into two groups: static-scene dependent and static-scene independent, and they are further classified based on input frames into multi-frame based and single-frame based.  We employ codes and pre-train models of TMT and TurbRecon  designed for static scenes to ensure a fair comparison. All methods incorporate deblurring effects, with NDIR and NDL using their default deblurring approach \cite{shan2008high}.

\subsection{Qualitative and Quantitative Evaluation}\label{sec5.2}

\noindent\textbf{Qualitative Evaluation on Real Turbulence.} In Fig. \ref{exp_real}, we compare with the existing methods on the RLR-AT. Single-frame based methods TurbNet \cite{mao2022single} and PiRN \cite{jaiswal2023physics} struggle with distortions as they lack modeling of temporal information for turbulence. The results of supervised-based methods like TSR-WGAN \cite{jin2021neutralizing} and TMT \cite{zhang2022imaging} still exhibit  distortions due to the domain gap. CLEAR \cite{anantrasirichai2013atmospheric}, SG \cite{lou2013video} and NDL \cite{zhu2012removing} continually produce artifacts or distortions due to unsuitable design for long-range turbulence. Albeit TurbRecon \cite{mao2020image} can acquire results with comparable quality, the results still encounter misalignments.
In comparison, CDSP consistently achieves more pleasing results at various distances, effectively addressing severe distortions while preserving details.  We also conduct a comparison on hot-air turbulence, which are shown in the appendix.

\noindent\textbf{Quantitative Evaluation on Synthetic Turbulence.} We further evaluate the performance of CDSP and other methods on synthetic turbulence in Table \ref{comparison}. It is observed that most multi-frame based methods perform better than single-frame based methods, since they take into consideration the temporal information of turbulence. CDSP and TurbRecon achieve the best performance in their respective categories. Note that as the distance increases, CDSP outperforms existing SOTA methods even more significantly,  further revealing the superiority of CDSP for long-range turbulence with severe distortions. 

\begin{table*}[t]
  \centering
  \setlength{\abovecaptionskip}{2pt}
  \setlength{\belowcaptionskip}{2pt}
  \setlength{\arrayrulewidth}{0.8pt}
  \begin{adjustbox}{width=0.93\columnwidth}
  \begin{minipage}[t]{0.229\textwidth}\normalsize

    \caption{Ablation study of FRF and SLRTR.}
    \centering
    \setlength\tabcolsep{0.3pt}
    \begin{adjustbox}{width=1\columnwidth}
    \renewcommand\arraystretch{1.80}
    \begin{tabular}{cc|cc}
      \hline
				FRF & SLRTR&PSNR  &SSIM   \\
        \hline
      \ding{55}&\ding{55}& 22.62& 0.8051 \\
      \ding{55}&\ding{51} &23.02& 0.8034 \\
				\ding{51}&\ding{55}&23.53 &0.8154 \\
        \ding{51}&\ding{51} &\textbf{23.96} & \textbf{0.8385}\\
        \hline
    \end{tabular}
  \end{adjustbox}
  \label{xiaorong}
  \end{minipage}
  \hfil
  \begin{minipage}[t]{0.4\textwidth}  \LARGE
    \caption{Effectiveness of FRF when embedded into existing methods compared to others. }
    \centering
    \captionsetup{position=bottom} 
    \renewcommand\arraystretch{1.2}
    \setlength{\arrayrulewidth}{1.8pt}
    \setlength\tabcolsep{3pt}
    \begin{adjustbox}{width=1\columnwidth}
    \begin{tabular}{c|c|cccc|cc>{\columncolor{gray!0}}c}
    \hline
    \multirow{3}*{Methods}&\multirow{3}*{Metric}&\multicolumn{4}{c|}{Reference frame} &\multirow{3}*{\textbf{Gain}}\\ \cline{3-6}
    & & &\makecell{Temp avg \\ \cite{zhu2012removing}}     &\makecell{Non-local avg \\ \cite{mao2020image}}     &\makecell{\textbf{FRF} \\ \textbf{(Ours)}} \\ 
    \hline

    \cellcolor{white}\multirow{2}*{\makecell{NDL \\ \cite{zhu2012removing}}}&PSNR    & &22.95 &22.88  &\textbf{23.29} &\textbf{0.34}\\ 
    &SSIM  & &0.7698 &0.7650  &\textbf{0.7819} &\textbf{0.0121}  \\ 
    \hline

    \cellcolor{white}\multirow{2}*{\makecell{TurbRecon \\ \cite{mao2020image}}}&PSNR   & &24.33 &24.14  &\textbf{24.79} &\textbf{0.65} \\ 
    &SSIM & &0.8122 &0.8082  &\textbf{0.8231} &\textbf{0.0149}   \\ 
    \hline

    \cellcolor{white}\multirow{2}*{\makecell{\textbf{CDSP} \\ \textbf{(Ours)}}}&PSNR   & &\textbf{25.25} &\textbf{24.68}  &\textbf{25.41} &\diagbox{}{}\\ 
    &SSIM & &\textbf{0.8391} &\textbf{0.8331}  &\textbf{0.8462} &\textbf{\diagbox{}{} }  \\ 
    \hline
    \end{tabular}
    \end{adjustbox}
    \label{FRF}
  \end{minipage}
  \hfill
  \begin{minipage}[t]{0.3\textwidth}  \normalsize
    \caption{Boosting performance of high-level text recognition task.}
    \setlength\tabcolsep{0.3pt}
    \begin{adjustbox}{width=1\columnwidth}
    \renewcommand\arraystretch{1.02}
    \begin{tabular}{c|ccc}
      \hline
				Methods & CRNN &ASTERN  &DAN  \\
        \hline
      Distorted&0.2553& 0.3475& 0.3759 \\
      TurbNet& 0.2057 & 0.3546& 0.3617 \\
			PiRN&0.2695&0.3546&0.3758 \\
        SG&0.2340&0.3971 &0.4042\\
        NDL&0.4539 &0.4894 & 0.4681\\
        TMT&0.4040&0.4893& 0.4964\\
        TurbRecon&0.4397 &0.4965&0.4894 \\
        \hline
          \textbf{CDSP}&\textbf{0.5248} &\textbf{0.5319} & \textbf{0.5532}\\
          \hline
    \end{tabular}
  \end{adjustbox}
  \label{Text}
  \end{minipage}

\end{adjustbox}
\end{table*}

\begin{figure*}[t]
  \centering
  \captionsetup{skip=0.8mm} 
  \centering
     \includegraphics[width=0.93\linewidth]{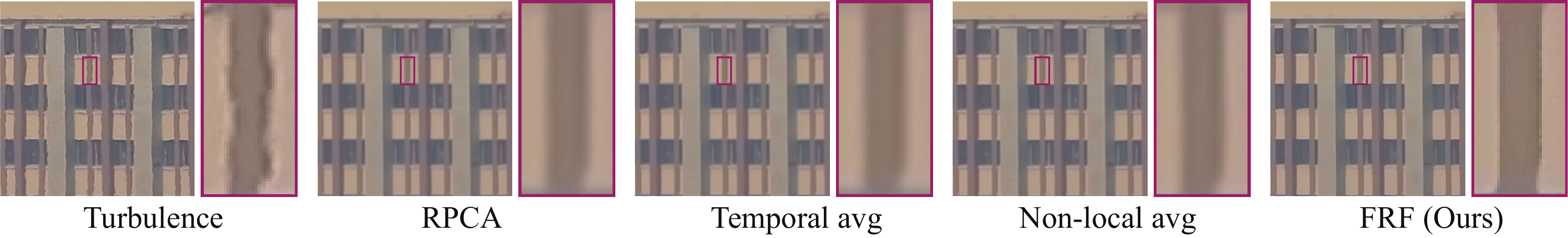}
  \caption{Visual comparison of FRF with other reference frame.}
  \label{REF_FIG}
\end{figure*}

\subsection{Ablation and Discussion}\label{sec5.3}

\noindent\textbf{How dose FRF Facilitate Severe Distortion Correction?}
We study the importance of FRF for distortion correction. As shown in Fig. \ref{referenceframe}(a) earlier, due to the superior quality  and sharp edges of FRF, better registration can be achieved, greatly reducing the burden of refinement.  We remove the FRF-based registration (FRF), and directly employ SLRTR  to handle severe distortions, placing a heavy burden on SLRTR. Table \ref{xiaorong} shows that the removal of FRF leads to a noticeable performance drop. Moreover, the result without FRF in Fig. \ref{EFF_fig}(a) encounters severe detail loss. Figure \ref{EFF_fig}(b) and (c) show  the visualization and distribution of residuals from SLRTR decomposition. It is evident that the residual without FRF contains more lost details. This indicates that FRF is indispensable in reducing the burden of refinement and preventing severe details loss.  

\noindent\textbf{How dose SLRTR Improve Severe Distortion Correction?}
We aim to emphasize the necessity of SLRTR for distortion correction. We remove the SLRTR and directly utilize FRF-based registration. However, achieving perfect pixel-level registration is impossible, since there exist severe distortions in long-range turbulence. Table \ref{xiaorong} shows that the method encounters a obvious performance drop without SLRTR, and the result without SLRTR in Fig. \ref{EFF_fig}(a) still exists  misalignments, indicating that SLRTR is necessary for refining registration errors. 

\noindent\textbf{Effectiveness of Subspace and Self-similarity.} 
Then we aim to illustrate the effectiveness of subspace and self-similarity to the proposed SLRTR. The subspace is utilized to characterize the global low-rank property along the temporal dimension. In Fig. \ref{EFF_SLRTR}(b), the result without subspace still exists distortions, indicating that relying solely on the  prior of spatial self-similarity is insufficient to characterize the properties of the temporal dimension. The non-local prior is employed to explore the self-similarity of the spatial dimension. Figure \ref{EFF_SLRTR}(c) shows the result without self-similarity, which suffers from unexpected details loss, implying that relying only on the temporal information is not enough.

\noindent\textbf{Complementarity between SLRTR and FRF.}
We further discuss how does FRF and SLRTR complement each other in Fig. \ref{EFF_fig}(a). On one hand, FRF-based registration could notably mitigate distortions with fewer corruptions, greatly reducing the burden of SLRTR. On the other hand, the SLRTR could effectively refine the residual errors unavoidably left by FRF-based registration. FRF and SLRTR complement  to each other to better remove the severe distortion.

\noindent\textbf{Effectiveness of Frequency-aware Reference Frame.}
To further illustrate the Effectiveness of FRF, we embed existing reference frames (Temp Avg \cite{zhu2012removing}, Non-local average \cite{mao2020image}) and proposed FRF into existing frameworks: NDL \cite{zhu2012removing}, TurbRecon \cite{mao2020image} and proposed CDSP on 5 Km synthetic turbulence. Table \ref{FRF} shows that existing methods obtain the highest PSNR/SSIM after integrating FRF, revealing the effectiveness of FRF. We also visualize the comparison between FRF and other reference frames in Fig. \ref{REF_FIG}. It is observed that FRF possesses superior quality and sharper edge, further demonstrating the reliability of FRF.


\noindent\textbf{Promotion for Downstream Recognition.} We further evaluate the turbulence mitigating methods on text recognition using TurbuText \cite{Textdataset}. We apply three text recognition methods (CRNN \cite{shi2016end}, ASTERN \cite{shi2018aster}, DAN \cite{wang2020decoupled}) on restoration results and report the accuracies in Table \ref{Text}. CDSP consistently improves the recognition performance for all recognition methods.

\begin{figure*}[t]
  \captionsetup{skip=0.8mm} 
  \centering
     \includegraphics[width=0.90\linewidth]{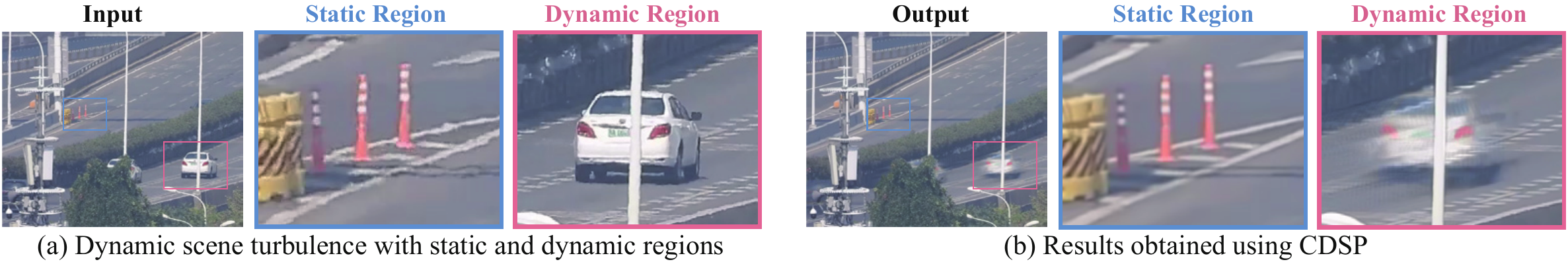}
  \caption{Limitation of the proposed CDSP. (a) Dynamic scene turbulence. (b) Result of CDSP. CDSP can effectively handle static region but fails to process dynamic regions. }
  \label{limitation}
\end{figure*}

\noindent\textbf{Limitation.} The proposed method could effectively handle static scene turbulence, but dynamic scenes (\textit{e.g.} scene contains moving objects or camera shake) are more complex due to the coupling of turbulence, object and camera motion. Figure \ref{limitation} shows the results of dynamic scene. Though static regions are well-processed, dynamic regions suffer from severe trailing since CDSP does not model object motion. We will address  dynamic scene turbulence in future work.

\section{Conclusion}
Our work focuses on long-range turbulence mitigation. We construct a long-range turbulence dataset (RLR-AT).  We propose a coarse-to-fine framework for long-range turbulence mitigation,  which cooperates the dynamic and the static priors. We propose a frequency-aware reference frame for better registration. We  propose a low-rank tensor refinement model to refine the registration error with details preserving.  Extensive experiments demonstrate the proposed method outperforms SOTA methods on different datasets.

\section*{Acknowledgements}
This work was supported by the National Natural Science Foundation of China under Grant
62371203. The computation is completed in the HPC Platform of Huazhong University of Science and Technology.


%
%
\bibliographystyle{splncs04}
\bibliography{main_v3}

\end{document}